\documentclass[11pt]{article}

\usepackage[margin=1in]{geometry}
\usepackage[T1]{fontenc}
\usepackage{lmodern}
\usepackage{microtype}
\usepackage{amsmath,amssymb}
\usepackage{booktabs}
\usepackage{array}
\usepackage{graphicx}
\usepackage{xcolor}
\usepackage{enumitem}
\usepackage{caption}
\usepackage{siunitx}
\usepackage{pgfplots}
\pgfplotsset{compat=1.17}
\usepackage[colorlinks=true,linkcolor=blue!55!black,citecolor=blue!55!black,urlcolor=blue!55!black]{hyperref}
\usepackage{xurl}   

\DeclareRobustCommand{\paritok}{\textsc{Paritok-4B}}
\DeclareRobustCommand{\code}[1]{\texttt{\frenchspacing #1}}  

\title{\textbf{Paritok-4B: Intent-Conditioned Context\\ Compression for Coding Agents}}

\author{Jiayu Shi \qquad Luzhuo Chen\\[2pt]
\small Paritok\\
\small \texttt{paritok9@gmail.com}}

\date{\today}

\begin{document}
\maketitle

\begin{abstract}
Coding agents re-send large file reads and tool outputs to a frontier LLM on every turn, and this context dominates their token bill. Prompt-compression models can shrink that context, but general-purpose compressors are trained on prose and are not well suited to code: they paraphrase identifiers and drop the exact spans an agent needs to edit. We present \paritok{}, a 4B-parameter LoRA compressor trained specifically for coding-agent trajectories under two design commitments: (i) \emph{extractive}, so the model selects spans rather than rewriting them---measured over the training corpus, 96.0\% of emitted identifiers, paths, and numbers already appear in the input (98.3\% outside the one kind we rewrite by design), and the same audit run on held-out SWE-bench Lite output gives 96.2\% over 212{,}506 emitted tokens; and (ii) \emph{intent-conditioned}, so the compressor is told the agent's current task---which we measure to act chiefly \emph{inside} a retained segment, selecting which lines survive (retained lines are $+0.067$ more intent-relevant than removed ones, paired 95\% CI $[+0.056,+0.078]$) rather than changing how much is retained. The compressor operates \emph{per segment}: a gateway splits the agent request into typed segments and compresses each independently, which makes the unit of work small, parallelizable, and individually recoverable. We build the training set by distilling a \code{gpt-4.1-mini} teacher over 67{,}074 real OpenHands agent trajectories through a five-stage funnel, yielding a 45K-segment distillation pool and \num{40606} validated examples, and fine-tune a Qwen3-4B backbone with LoRA, selected over 3B and 7B code-pretrained alternatives under a matched protocol. On an out-of-distribution holdout the released checkpoint emits well-formed output on 100\% of segments, compresses to 23.7\% of input tokens, and---at that budget---shows no degradation in must-keep identifier retention relative to the teacher it was distilled from (0.385 vs.\ 0.287 on segments both kept; the paired difference is directionally favorable but not significant at $n{=}39$). End-to-end on all 300 SWE-bench Lite instances, \paritok{} compresses agent context to \textbf{25.7\%} of its size---$2.0\times$ harder than a \code{gpt-4.1-mini} compressor (50.2\%) and $2.4\times$ harder than \code{gpt-5} (61.9\%)---while retaining \textbf{86.5\%} of uncompressed single-shot solve quality, on par with the \code{gpt-4.1-mini} compressor at less than half the tokens; run instead in the in-distribution \code{cat -n} regime that real agents produce, the same 300 instances compress slightly less (27.8\%) and retain more (\textbf{89.3\%}), where the paired comparison is the informative one: 30 instances are solved only uncompressed and 17 only compressed, an exact McNemar $p{=}0.079$---at this sample size, compressing context to roughly a quarter of its size does \emph{not} significantly reduce the solve rate. We are explicit (\S\ref{sec:e2e}) that this harness measures comprehension under compression, not end-to-end agent cost. The model is a 264\,MB LoRA adapter that self-hosts on a single 24\,GB GPU with no per-token compressor fee---which, at list prices, is what decides the economics: we show (\S\ref{sec:cost}) that \code{gpt-5} used as a compressor is \emph{net-negative}, costing more to run than the downstream tokens it saves. Weights, data pipeline, and evaluation scripts---including the extractiveness audit---are open (Apache~2.0).
\end{abstract}

\section{Introduction}

An autonomous coding agent (Claude Code, Cursor, Codex, OpenHands~\cite{openhands}) solves a task through many turns of reading files, running commands, and editing code~\cite{react}. Each turn re-sends the accumulated context---file reads, command output, history---to a frontier LLM, so this input, not the model's output, dominates the token bill. Compressing that context with a small model before it reaches the expensive LLM is an attractive lever, and prompt-compression research shows heavy compression can preserve task quality~\cite{llmlingua,llmlingua2,selectivecontext}.

But a compressor for \emph{coding agents} faces constraints that general prose compressors were never built for. First, an agent edits code by exact string match: if the compressor paraphrases a function signature or renames a variable, the downstream edit fails. Second, the value of a segment is not intrinsic---it depends on what the agent is currently doing; the function the agent is about to modify must survive even if it looks unremarkable. Third, agent context is strongly heterogeneous: a \code{cat -n} file read, a pytest traceback, an \code{ls} listing, and a chain-of-thought block have nothing in common, and a single uniform compression ratio is the wrong tool for all four.

We address these with \paritok{}, a 4B LoRA compressor trained on real coding-agent trajectories under two commitments:

\begin{enumerate}[leftmargin=1.4em,itemsep=2pt]
  \item \textbf{Extractive.} For code and tool output the model selects spans rather than rewriting them, drawing on a closed, fixed vocabulary of structural markers for what it removes (\S\ref{sec:extractive}). Identifiers, paths, and error strings are preserved by copying rather than by the model's judgment about how to reword them---96.0\% of emitted identifier-like tokens are already present in the input, and 96.2\% on held-out SWE-bench Lite output. We audit this claim rather than assert it, and report where it leaks.
  \item \textbf{Intent-conditioned.} The compressor receives the agent's current task/query and is trained to keep the entities that task names (its highest-priority rule), so ``what matters'' is defined relative to the agent's live intent rather than statically.
\end{enumerate}

A third mechanism, a four-level importance label (L0--L3) carrying a per-level compression budget, is part of the pipeline but is \emph{not} among the commitments we claim: \S\ref{sec:levels} shows the distilled targets realize only two effective bands rather than four, so we report it as a design that did not deliver what it was meant to.

Our contributions are: (1) a reproducible \textbf{five-stage data pipeline} (\S\ref{sec:data}) that turns 67{,}074 raw OpenHands trajectories into a 45K-segment distillation pool and \num{40606} teacher-validated compression examples, with intent-conditioned, level-labeled, must-keep-annotated targets; (2) a \textbf{training recipe reported with its dead ends} (\S\ref{sec:train}): a matched three-backbone comparison behind the 4B choice, a drop-supervision weighting that backfired and why we shipped without it, and a deployed checkpoint chosen by an out-of-distribution sweep rather than by training loss---including why the loss-optimal checkpoint was \emph{not} the one shipped; (3) an \textbf{intrinsic evaluation} (\S\ref{sec:intrinsic}) on an OOD holdout, reported with confidence intervals, including the finding that the student's must-keep identifier retention is not below its own teacher's at a comparable budget; and (4) an \textbf{end-to-end evaluation} (\S\ref{sec:e2e}) showing \paritok{} reaches a 25.7\% compression rate---roughly twice as aggressive as strong GPT compressors---at comparable single-shot solve quality, reported in both the raw-source and the in-distribution line-numbered regime---where the trade we predicted (harder compression at equal quality) is not the one we measured (slightly softer compression at higher quality, and no paired degradation resolvable at 300 instances)---with an explicit account of what the harness does and does not measure.

\section{Related Work}

\paragraph{Prompt compression.} LLMLingua~\cite{llmlingua} and Selective Context~\cite{selectivecontext} drop low-information tokens using a small language model's perplexity signal; LLMLingua-2~\cite{llmlingua2} distills a GPT-4 teacher into a token-classification compressor, making compression a supervised extractive task. \paritok{} inherits the distillation-into-a-small-model structure of LLMLingua-2 but differs in three ways: the unit is a typed agent segment rather than a prose passage, the target is conditioned on an explicit task query and an assigned importance level rather than being task-agnostic, and the training corpus is agent trajectories rather than documents. Task-agnostic compressors are attractive because one model serves every downstream use; we argue the opposite trade for coding agents, where the agent's current intent is available for free at compression time and is the single most informative feature for deciding what to keep.

We do not report a head-to-head against LLMLingua-2, and we would rather say why than leave the omission unexplained. The two systems do not accept the same input: LLMLingua-2 compresses a passage to a target ratio with no query, whereas \paritok{} is invoked per typed segment with an intent string and a level tag, and a third of its decisions are whole-segment drops that a token-classification compressor has no way to express. Any single protocol we could run both under would either strip \paritok{} of the inputs it was trained on or ask LLMLingua-2 for a decision it was not built to make, and the resulting number would say more about the harness than about either system. The comparison worth running is the end-to-end one---same context, same downstream agent, measure tokens and solve rate---and it is future work rather than a claim we make here.

\paragraph{Coding agents and their benchmarks.} SWE-bench~\cite{swebench} established issue-resolution on real repositories as the standard evaluation, and OpenHands~\cite{openhands} is a widely used open scaffold whose trajectories are published at scale by SWE-Gym~\cite{swegym} and SWE-rebench~\cite{swerebench}. These trajectory corpora are what make supervised training of an agent-specific compressor possible: they supply the actual distribution of file reads, tool results, and histories that a deployed compressor sees, which differs sharply from raw source files (agents read code through \code{cat -n}, with line-number framing and tool wrappers).

\section{Task Formulation}
\label{sec:task}

We frame context compression as \emph{intent-conditioned extractive selection, applied one segment at a time}. A coding-agent request is a sequence of segments $s_1,\dots,s_n$ (a system prompt, the user's task, prior file reads, command outputs, edits, and reasoning). A gateway performs segmentation, kind classification, and level labelling; the model is then invoked once per compressible segment with exactly two inputs---the agent's current task $q$, and the segment itself tagged with its kind and level---and returns that segment's compressed form. Notably the model receives no numeric token budget: the per-level target ratios live only as a static table in the system prompt, which the model must associate with the level tag in the segment header. \S\ref{sec:levels} shows this indirection did not survive training.

This per-segment decomposition is a deliberate design choice, not an implementation detail. It keeps each model call short (median prompt $\approx$4K tokens) so a 4B model with a 16K window suffices; it lets segments be compressed in parallel; it makes compression incremental across turns, since unchanged segments need not be recompressed; and it makes each unit individually recoverable, because a compressed segment maps back to exactly one original.

Each segment carries a \emph{kind} and an importance \emph{level}. The kinds are \code{file\_read}, \code{bash\_command}, \code{log\_output}, \code{tool\_result}, \code{file\_operation}, \code{directory\_listing}, \code{assistant\_thinking}, and \code{meta\_action}, alongside the protected \code{system} and \code{user} messages. The levels are:
\begin{itemize}[leftmargin=1.4em,itemsep=1pt]
  \item \textbf{L0} --- protected: system prompt, current user task, most-recent tool result (intended budget $\le 0.50$).
  \item \textbf{L1} --- recent reads and all actions, edits and commands (intended $\le 0.35$).
  \item \textbf{L2} --- mid-history reads and reasoning (intended $\le 0.25$).
  \item \textbf{L3} --- stale context: superseded file re-reads, ancient turns (intended $\le 0.20$).
\end{itemize}
We say \emph{intended} because the realized targets do not honour this four-way split---they collapse into two bands, protected/recent against stale, and \S\ref{sec:levels} reports the measurement. The asymmetry the design was after does exist, at half the resolution it was specified with. Dropping a segment entirely is a first-class action: the model emits an empty body, which is the correct output for unrelated helpers, build noise, and superseded re-reads. Output structure mirrors the input: each segment is emitted inside a \code{[SEG id=$s_k$ kind=$\dots$ level=$\dots$]\dots[/SEG]} marker reusing the input's \code{seg\_id}, so the gateway can map every compressed span back to its origin and recover the untouched bytes on demand.

\subsection{What ``extractive'' means here, precisely}
\label{sec:extractive}

We use \emph{extractive} in the summarization sense---the model selects and deletes rather than re-generating---and we state the boundaries explicitly, because the guarantee is only as strong as its exceptions.

\emph{Copied verbatim:} all retained code lines, identifiers, file paths, line numbers, imports, error classes, exact error-message text, shell commands, and the \code{old\_str}/\code{new\_str}/\code{file\_text} payloads of edits to project source, which are preserved in full without truncation.

\emph{A closed set of structural markers} may replace deleted content (Table~\ref{tab:markers}). These are structural pointers with a fixed grammar; they carry counts and names taken from the input, and the training rules forbid a body consisting only of markers. The set is closed in the strong sense: it is enumerated in the system prompt, and a marker outside it is a format violation.

\begin{table}[t]
\centering
\caption{The complete marker vocabulary. These are the only structural placeholders the system prompt sanctions; a marker outside this set is a format violation. Table~\ref{tab:extractive} measures how much of the output is copied rather than marked or generated.}
\label{tab:markers}
\small
\begin{tabular}{ll}
\toprule
Marker & Replaces \\
\midrule
\code{[file: basename.py]}            & the \code{cat -n} framing of a file read \\
\code{[body: N lines]}                & an elided function or class body \\
\code{[lines L1-L2: fnA / fnB -- note]} & a run of collapsed adjacent functions \\
\code{[imports: A, B, C]}             & a collapsed import block \\
\code{[N more matches in <file>]}     & truncated grep output \\
\code{[<line> \texttimes{} N]}         & a log line repeated $N$ times \\
\code{[N lines unchanged]}            & an elided diff tail \\
\code{[N lines elided]}               & generic long-output truncation \\
\code{[N tests collected]}            & pytest collection output \\
\code{[N more entries]}               & a truncated directory listing \\
\code{[plan: T1 -- status; \dots]}    & a compacted task-tracker plan \\
\bottomrule
\end{tabular}
\end{table}

\emph{Bounded rewriting} is permitted in a few places by rule: long string literals inside \code{raise}\slash\code{warn}\slash\code{assert} and logging calls may be abbreviated while keeping the exception type and any format specifiers; \code{assistant\_thinking} is reduced to a single $\le$200-character sentence stating the decision the agent reached, or dropped if it reached none; \code{meta\_action} plans become a $\le$300-character title-plus-status list; a segment that is entirely mid-docstring is re-emitted in a compact spec form; and a tool call arriving as JSON is unwrapped into a compact template---\code{view <path>}, or \code{replace <path>} followed by the old and new strings---whose payload is copied but whose framing is regenerated.

\paragraph{How extractive is it, measured.} ``Extractive'' is a design commitment, and commitments should be audited rather than asserted, so we measure it over the full distilled corpus (Table~\ref{tab:extractive}). Two views disagree in an informative way. At the \emph{line} level, 82.3\% of emitted lines are byte-identical to a span of the input, 3.2\% are closed-vocabulary markers, and 14.5\% are newly generated---but that last figure is dominated by the tool-call unwrapping above, which rewrites the framing of a line whose payload is copied, and by \code{assistant\_thinking}, which is abstractive by rule. At the \emph{token} level---identifiers, dotted paths, and numbers, the tokens whose invention would actually break an agent---\textbf{96.0\%} of emitted tokens already appear in the input, rising to \textbf{98.3\%} once \code{assistant\_thinking} is excluded, and to 97.9\% on \code{file\_read} and 99.3\% on \code{log\_output}.

\paragraph{And it holds off the training distribution.} Measuring extractiveness on the corpus the model was fit to is the weaker half of the question; the stronger half is whether the copy behavior survives on data it never saw. We therefore ran the identical measure over the model's output on all 300 SWE-bench Lite instances of \S\ref{sec:e2e}---64{,}843 emitted lines and 212{,}506 emitted identifier-like tokens, none of it in training. The two numbers land where the corpus predicts: \textbf{92.2\%} of lines are byte-identical to a span of the input (0.4\% markers, 7.5\% novel) and \textbf{96.2\%} of emitted identifiers, paths, and numbers already appear in the input, against 88.6\% and 97.9\% for \code{file\_read} in Table~\ref{tab:extractive}, which is the matching kind. Per instance the median token-copy rate is 97.0\% and 213 of 300 instances sit at or above 95\%. The low tail is an artifact of near-total drop rather than of invention: the five lowest-scoring instances emit 8, 8, 52, 129, and 8 identifier-like tokens in total, so their ratios are taken over a handful of surviving tokens in outputs compressed to roughly 1\% of input. We note the honest limit of this check---SWE-bench Lite instances are held out end to end, but its repositories are popular enough that some also appear among the SWE-rebench and SWE-Gym trajectories the corpus was distilled from, so this is unseen-instance rather than unseen-repository generalization.

We take the residual seriously rather than rounding it away. It is not measurement noise: the teacher occasionally synthesizes a fully-qualified name the input only partially contained (an input reading \code{Exception type: FailedParse} becoming \code{tatsu.exceptions.FailedParse:}), rewords a comment, or restructures a multi-line expression, and the student learned these along with everything else. So the accurate statement is that \paritok{} is \emph{substantially} extractive---copy-first by construction and 96--98\% copied in practice---and not that it is provably incapable of emitting a token the input lacked. The property an editing agent actually depends on is narrower and does hold by rule: retained code lines and error strings are copied, so an exact-match edit against the compressed context behaves as it would against the original. \S\ref{sec:limits} carries the operational safeguard this implies.

\begin{table}[t]
\centering
\caption{Measured extractiveness over the full distilled corpus (236{,}152 emitted lines), by segment kind. \emph{Line} columns are strict and penalize reformatting; \emph{token copy} is the fraction of emitted identifiers, dotted paths, and numbers that already appear in the input. Reproduced by \code{eval/extractiveness.py}.}
\label{tab:extractive}
\small
\begin{tabular}{lrrrr}
\toprule
Kind & Line verbatim & Marker & Line novel & Token copy \\
\midrule
\code{file\_read}          & 88.6\% & 5.1\% & 6.3\%  & 97.9\% \\
\code{log\_output}         & 94.6\% & 1.0\% & 4.4\%  & 99.3\% \\
\code{file\_operation}     & 55.4\% & 0.0\% & 44.6\% & 98.0\% \\
\code{directory\_listing}  & 93.4\% & 4.2\% & 2.4\%  & 95.8\% \\
\code{tool\_result}        & 90.7\% & 1.2\% & 8.1\%  & 99.1\% \\
\code{bash\_command}       & 46.5\% & 0.0\% & 53.5\% & 98.1\% \\
\code{assistant\_thinking} & 0.0\%  & 0.0\% & 100.0\% & 71.8\% \\
\code{meta\_action}        & 7.7\%  & 84.5\% & 7.8\% & 98.9\% \\
\midrule
\textbf{All}                       & 82.3\% & 3.2\% & 14.5\% & \textbf{96.0\%} \\
All except \code{assistant\_thinking} & --- & --- & --- & \textbf{98.3\%} \\
\bottomrule
\end{tabular}
\end{table}

\section{Data Pipeline}
\label{sec:data}

The training signal is distilled from a teacher over real agent trajectories through a five-stage funnel (Table~\ref{tab:funnel}). Every stage is scripted and re-runnable.

\begin{table}[t]
\centering
\caption{Data funnel. Stages 1--5 operate on \emph{turn-level} samples; stage 6 extracts individual \emph{segments} from a stratified subset of the 80K turn pool and distills them, since the deployed model compresses one segment per call (\S\ref{sec:task}). The two units should not be compared directly.}
\label{tab:funnel}
\small
\begin{tabular}{llll}
\toprule
Stage & Script & Unit & Scale \\
\midrule
Source trajectories (OpenHands) & \code{01\_download} & trajectory & \num{67074} \\
Turn-level samples (segmented)  & \code{02\_parse}    & turn & \num{469518} \\
Filtered (compressible, real action) & \code{03\_filter} & turn & \num{423358} \\
Labeled (levels + must-keep)    & \code{04\_label}    & turn & L0--L3 + spans \\
Finalized pool (traj-level split) & \code{05\_finalize} & turn & 80K train / 4K val \\
\midrule
Per-segment distillation pool   & \code{06\_distill}  & segment & 45{,}000 \\
\quad \code{file\_read}         &                     & segment & 10{,}000 \\
\quad other kinds               &                     & segment & 35{,}000 \\
Teacher-distilled + validated   & \code{06\_distill}  & segment & \textbf{\num{40606}} \\
\quad \code{file\_read}         &                     & segment & \num{9830} \\
\quad other kinds               &                     & segment & \num{30776} \\
\bottomrule
\end{tabular}
\end{table}

\paragraph{Stage 1 --- source.} We download OpenHands agent trajectories from SWE-rebench~\cite{swerebench} and SWE-Gym~\cite{swegym} ($\sim$67K trajectories of real GitHub-issue resolution), keeping SWE-bench Lite only as a held-out evaluation set, never for training.

\paragraph{Stage 2 --- segmentation.} Each trajectory is split at assistant decision points (capped at 8 per trajectory, uniformly subsampled) into (history $\to$ next-action) samples. Every history message becomes a segment with a heuristically classified kind; \code{tool\_result} blocks are re-classified by content (e.g.\ \code{cat -n} output $\to$ \code{file\_read}). Crucially, file reads retain their literal line-numbered (\code{cat -n}) framing, matching how agents actually read files---a compressor trained on raw file text is out-of-distribution on real agent input.

\paragraph{Stage 3 --- filtering.} We keep only samples whose target is a real action (edit/command, not a \code{think}/\code{finish} meta-step) and that contain at least one compressible segment ($\ge$\,1000 tokens). Over-long segments ($>$\,4000 tokens) are split at line boundaries; requests over 32K tokens are middle-truncated---the largest middle segments are dropped first while the leading 30\% and trailing 50\% of \emph{segments} are retained---rather than being discarded outright.

\paragraph{Stage 4 --- labeling.} Without any LLM, each segment is assigned its L0--L3 level from kind, relative position, and recency, and a must-keep span set (paths, identifiers, error classes, line numbers, code keywords) is extracted. A stale-file detector marks superseded re-reads as L3.

\paragraph{Stage 5 --- pooling.} The turn-level pool is capped at 80K train / 4K val. The split is trajectory-level (5\% val, seed 42) so no trajectory leaks across it, and the pool is stratified by (length bucket, action type, resolved-flag) via two-pass reservoir sampling over file offsets.

\paragraph{Stage 6 --- distillation and validation.} Individual segments are extracted from the pool and distilled separately, with two teacher prompts---one for \code{file\_read}, one for all other kinds---since their compression rules differ substantially. Targets are produced by a \code{gpt-4.1-mini-2025-04-14} teacher via the OpenAI Batch API ($T{=}0$), prompted with the agent's intent, the segment's level, and the per-level budget, and instructed to preserve must-keep spans verbatim, condense L2 to one-liners, and drop L3. Every teacher output is validated and rejected if it exceeds $1.3\times$ or falls below $0.2\times$ the budget; if it is near-identical to the input (length ratio $>0.65$, tightened from an initial $0.85$ after weak compression slipped through); if the sample contains $\ge$2 ``lazy'' file-read segments (kept nearly whole); if an L0/L1 segment marked as shrunk retains $<$70\% of its must-keep spans; or if it hallucinates a preamble or wraps the code in prose. A rollout gate halts the run if the validation pass-rate falls below 70\%. Distillation ran in cost-staged batches (100 / 1{,}000 / 30{,}000 / 50{,}000 requests), and \num{40606} of the 45{,}000 candidates passed validation (90.2\%).

\paragraph{Teacher prompt hill-climbing.} The teacher prompt itself was tuned by a human-in-the-loop hill-climb against $\sim$20 hand-curated gold compressions across fifteen prompt versions (v5--v15). Early versions systematically \emph{under}-compressed (keeping docstrings and redundant imports, ratios near 0.5--0.9 vs.\ gold 0.25--0.4); later versions added an explicit ``named entity from intent stays complete'' rule and a cap against over-folding (hiding $>$half of retained lines behind a \code{[body: N lines]} marker). We also ran a teacher A/B at v15 across \code{gpt-4.1-mini}, \code{gpt-5}, and \code{gpt-5.1} variants; \code{gpt-4.1-mini} was retained as the shipped teacher for its cost/quality balance, and preference pairs targeting the over-/under-folding failure modes were collected for possible future DPO~\cite{dpo} refinement.

\subsection{The level design did not survive distillation}
\label{sec:levels}

The L0--L3 scheme is the one part of the design we cannot claim worked, and we report it rather than quietly dropping it, because the labels remain visible in the released data and a reader would find this anyway.

Measured over all \num{28248} kept segments in the distilled pool, the realized compression ratio does not separate into four levels (Table~\ref{tab:levels}). Only the L1/L2 boundary is real: a random L1 segment is compressed harder than a random L2 segment just 28.0\% of the time, the expected direction. At the other two boundaries the levels are indistinguishable---$P(\mathrm{L0}<\mathrm{L1})=0.534$ and $P(\mathrm{L2}<\mathrm{L3})=0.456$, both within noise of the 0.5 that means no separation, with L0/L1 mildly inverted against the intended ordering. The drop rates show the same two-band structure (15.5\% and 13.2\% against 38.3\% and 41.2\%). What the pipeline actually produces is a \emph{binary} distinction---protected-or-recent at roughly 0.40, stale at roughly 0.20---wearing four labels.

Two causes compound. The teacher was asked for four budgets and returned two bands, most visibly at L0, where it compresses to 0.379 against an instruction of 0.50. And the student was never given the budget as a number: it sees only \code{level=L2} in the segment header and must associate that tag with a row of a static table in its system prompt. An indirection that weak, supervised by targets that themselves do not separate, has nothing to teach. The student reproducing one policy across L0 and L1 is therefore the expected outcome, not a symptom of undertraining.

The fix is a data and interface change rather than more training: pass the per-segment token budget explicitly as a number, and derive it from the teacher's realized distribution instead of a hand-set table. We leave that to v2 and, in the meantime, make no level-awareness claim.

\begin{table}[t]
\centering
\caption{Realized compression by intended level, over all kept segments in the distilled pool. The right column is the probability that a random segment from this level is compressed harder than a random segment from the next; well below 0.5 means the boundary is real. Reproduced by \code{eval/design\_claims\_audit.py}.}
\label{tab:levels}
\small
\begin{tabular}{lccccc}
\toprule
Level & Intended & Median realized & Drop rate & $n$ & vs.\ next level \\
\midrule
L0 & $\le 0.50$ & 0.379 & 15.5\% & 1{,}500 & $0.534$ \\
L1 & $\le 0.35$ & 0.408 & 13.2\% & 4{,}665 & $\mathbf{0.280}$ \\
L2 & $\le 0.25$ & 0.198 & 38.3\% & 4{,}913 & $0.456$ \\
L3 & $\le 0.20$ & 0.133 & 41.2\% & 3{,}048 & --- \\
\bottomrule
\end{tabular}
\end{table}

\subsection{Where intent-conditioning actually acts}
\label{sec:intentaudit}

Intent-conditioning is the paper's title claim, so it deserves an audit, and the audit relocated it. It acts at a finer granularity than we first described, and barely at all where we first looked.

\paragraph{Not, mostly, at the segment level.} If the compressor filtered segments by relevance, intent overlap should predict the teacher's keep/drop decision. It predicts it weakly: as a lone ranker, the fraction of a segment's identifiers named in the intent reaches AUC 0.589 against that decision. Segment \emph{kind} alone reaches 0.762, kind and level together 0.838, and adding intent moves that only to 0.864. The reason is visible in the per-kind drop rates---\code{assistant\_thinking} 0.0\%, \code{bash\_command} 0.3\%, \code{meta\_action} 73.9\%---which are near-deterministic consequences of rules in the system prompt, not judgments about the task. Bucketing segments by intent overlap does show drop rates falling from 41.6\% to 16.5\%, but splitting within each \code{(kind, level)} cell shrinks that to 32.9\%~$\to$~28.0\%, with 13 of 20 cells in the expected direction. Most of the apparent segment-level effect is kind wearing intent's clothes.

\paragraph{But clearly at the line level, inside the segments it keeps.} The sharper test holds the segment fixed and asks which of its lines survived. For each retained \code{file\_read} we split the original into the lines the teacher kept and the lines it removed, and measure how much of each half's identifier set is named in the intent. Kept lines score 0.234, removed lines 0.167: a paired difference of \textbf{+0.067} with a bootstrap 95\% CI of $[+0.056, +0.078]$, holding in 64.5\% of segments ($n{=}785$). Because the comparison is within a single segment, it is automatically controlled for kind, level, repository, file identity, and length---every confound that contaminated the segment-level measurement.

\paragraph{This resolves an apparent contradiction.} Retained-segment compression \emph{ratio} is flat across intent buckets (0.303, 0.362, 0.364, 0.344, 0.297, 0.285), which read as evidence against intent-conditioning until the line-level result explained it: intent does not change \emph{how much} survives, it changes \emph{which lines} do. That is the behaviour one would want, and it is invisible to any metric that only counts tokens---including, until we corrected it, our own intent-sensitivity probe, which measured keep ratio alone and would have returned a false negative.

So the accurate statement of the mechanism is narrower and better evidenced than ``the compressor understands the task'': \emph{segment survival is governed mostly by kind and rule, and within a surviving segment the retained lines are selected for relevance to the entities the task names.} The remaining caveat is unchanged---this is measured on the teacher's targets, the necessary condition; confirming the student inherited it requires the probe of \S\ref{sec:levels}'s companion script, which we have not run.

\paragraph{Final format.} A training example is an OpenAI-style \code{messages} triple: a \emph{system} prompt defining the extractive compression engine, its per-kind rules, its drop criteria, and its ``do not execute / answer / preamble'' constraints; a \emph{user} message carrying the agent's intent and the single level-tagged \code{[SEG]} block to compress; and the teacher's validated compressed \code{[SEG]} as the \emph{assistant} target. Both system prompts are prefixed with \code{/no\_think} at training and inference time, because Qwen3-Instruct emits \code{<think>} blocks by default, which breaks the required output format.

\section{Training}
\label{sec:train}

We fine-tune \code{Qwen/Qwen3-4B-Instruct-2507}~\cite{qwen3} with LoRA~\cite{lora} (rank $r{=}32$, $\alpha{=}64$, dropout $0$) on the seven attention and MLP projections, using the Unsloth~\cite{unsloth} trainer and TRL's \code{SFTTrainer}~\cite{trl}. \S\ref{sec:backbone} explains why this backbone. Full hyperparameters are in Table~\ref{tab:hparams}. Training is bf16 (not QLoRA), single-GPU, with 8-bit AdamW and gradient checkpointing; the resulting adapter is 264\,MB.

\begin{table}[t]
\centering
\caption{SFT configuration. Effective batch $=$ per-device $2\times$ grad-accum $16$. The schedule was configured for 2 epochs (2{,}538 steps); the released adapter is the step-2{,}000 checkpoint (1.58 epochs), selected as described in \S\ref{sec:ckpt}.}
\label{tab:hparams}
\small
\begin{tabular}{ll}
\toprule
Base model & Qwen3-4B-Instruct-2507 \\
Adapter & LoRA $r{=}32$, $\alpha{=}64$, dropout $0$, bias none \\
Target modules & q,k,v,o,gate,up,down\_proj \\
Precision & bf16 (base not quantized) \\
Optimizer & 8-bit AdamW, wd $0.01$ \\
Learning rate & $1{\times}10^{-5}$, linear decay, warmup $0.1$ \\
Batch & $2\times16$ (effective 32) \\
Max sequence length & 16{,}384 \\
Configured schedule & 2 epochs / 2{,}538 steps \\
\textbf{Released checkpoint} & \textbf{step 2{,}000 (1.58 epochs)} \\
Tokens seen at release step & \num{300524776} \\
Checkpoint / eval interval & every 200 steps \\
Seed & 42 \\
Hardware & 1$\times$ H100 80\,GB (Unsloth, no DeepSpeed) \\
Adapter size & 264\,MB \\
\bottomrule
\end{tabular}
\end{table}

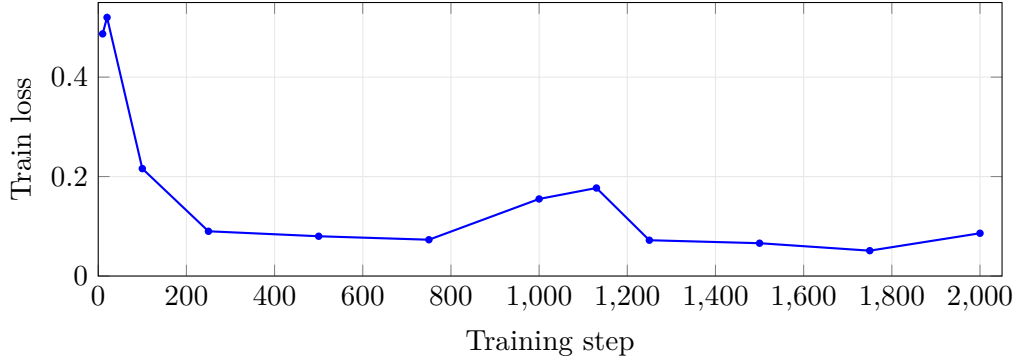
\begin{figure}[t]
\centering
\begin{tikzpicture}
\begin{axis}[
  width=0.82\linewidth, height=5.2cm,
  xlabel={Training step}, ylabel={Train loss},
  xmin=0, xmax=2050, ymin=0, ymax=0.55,
  grid=both, grid style={gray!18},
]
\addplot[thick, blue, mark=*, mark size=1pt] coordinates {
(10,0.487)(20,0.520)(100,0.216)(250,0.090)(500,0.080)(750,0.073)
(1000,0.155)(1130,0.177)(1250,0.072)(1500,0.066)(1750,0.051)(2000,0.086)
};
\end{axis}
\end{tikzpicture}
\caption{SFT loss: rapid convergence after the 10\% warmup, with one transient spike peaking at step 1{,}130, plateauing around $0.05$--$0.09$.}
\label{fig:loss}
\end{figure}

\paragraph{Loss.} Training loss drops sharply and plateaus below $0.09$ (Table~\ref{tab:loss}, Figure~\ref{fig:loss}), with token accuracy reaching $\sim$0.975. A single transient spike occurs around step 1{,}000--1{,}130 (peak $0.177$ at step 1{,}130) and recovers within roughly 100 steps; it corresponds to a stretch of unusually long \code{file\_operation} targets, where project-source edits must be reproduced verbatim in full.

\begin{table}[t]
\centering
\caption{SFT training-loss trajectory (logged every 10 steps; representative points).}
\label{tab:loss}
\small
\begin{tabular}{lrrrrrrrr}
\toprule
Step & 10 & 100 & 250 & 500 & 750 & 1{,}250 & 1{,}750 & 2{,}000 \\
\midrule
Loss & 0.487 & 0.216 & 0.090 & 0.080 & 0.073 & 0.072 & 0.051 & 0.086 \\
Tok.\ acc. & 0.928 & 0.951 & 0.967 & 0.964 & 0.968 & 0.976 & 0.978 & 0.975 \\
\bottomrule
\end{tabular}
\end{table}

\subsection{Backbone selection}
\label{sec:backbone}

The 4B backbone was chosen, not assumed. We ran the same SFT on three candidates (Table~\ref{tab:backbones}) under a deliberately matched protocol: identical training script and data, LoRA $r{=}32$/$\alpha{=}64$/dropout~0 on the same seven projections, learning rate $1{\times}10^{-5}$ with linear decay and 10\% warmup, effective batch 32, a 2-epoch schedule, a 16{,}384-token window, bf16 with the base unquantized, and seed 42. Only the base model varied. Per-device batch size had to differ---the 3B and 4B runs OOM in \code{F.cross\_entropy} above batch 2 at a 16K sequence with Qwen's 152K vocabulary, while the 7B run tolerates batch 8 on a 94--141\,GB card---but gradient accumulation was set to hold the effective batch at 32 in every run, so the optimization math is the same.

\begin{table}[t]
\centering
\caption{The three SFT backbones, trained under a matched protocol. Per-device batch differs only to fit memory; gradient accumulation holds the effective batch at 32 throughout.}
\label{tab:backbones}
\small
\begin{tabular}{llll}
\toprule
Backbone & Family & Per-device batch & Outcome \\
\midrule
\code{Qwen2.5-Coder-3B-Instruct} & code-pretrained & 2 (grad-accum 16) & not shipped \\
\textbf{\code{Qwen3-4B-Instruct-2507}} & general instruct & 2 (grad-accum 16) & \textbf{shipped as v1} \\
\code{Qwen2.5-Coder-7B-Instruct} & code-pretrained & 8 (grad-accum 4) & not shipped \\
\bottomrule
\end{tabular}
\end{table}

Two considerations decided it. The first is the deployment envelope, and it is the harder constraint. The product target is a compressor a team self-hosts next to its agent on a single commodity 24\,GB GPU, running at a 16K context on every turn. A 7B backbone does not fit that envelope with room for the KV cache at this context length, and the training runs show the same pressure from the other side: the 7B configuration only reaches a comfortable per-device batch on 94--141\,GB cards and must fall back to batch 4 at roughly 85\% VRAM on an 80\,GB card. A compressor that costs more to host than the tokens it saves defeats its own purpose, which makes the size ceiling a first-class design constraint rather than a budget detail.

The second is that the code-pretrained advantage did not materialize. The intuition favors \code{Qwen2.5-Coder} for a task whose inputs are source files and tracebacks, but this task is not code \emph{generation}---the model copies spans and decides what to discard, and the decision is driven by the natural-language intent string, not by the ability to write correct Python. The general-instruct Qwen3-4B follows the structured output contract and the intent-conditioning rule at least as reliably as the code-pretrained candidates, while sitting below the deployment ceiling. It carries one backbone-specific quirk, handled in the data format: Qwen3-Instruct emits \code{<think>} blocks by default, which breaks the required \code{[SEG]} output, so both system prompts are prefixed with \code{/no\_think} at training and inference time.

We are explicit about the limits of this comparison. The 3B and 7B runs predate a change in drop supervision---they used a drop-loss weight of 20, which we later found over-drops badly (\S\ref{sec:dropweight})---and the shipped 4B model was retrained at weight~1 afterwards. The three runs are therefore matched on optimization but not on drop supervision, and we release holdout artifacts only for the 4B run. This subsection is the rationale for the choice, not a benchmarked ablation; a clean three-way comparison under the final recipe is future work.

\subsection{Drop supervision: a weighting that backfired}
\label{sec:dropweight}

Dropping a segment is the highest-value action the model can take---it removes the whole segment---but it is the weakest training signal, because a drop target is roughly 10 tokens of empty \code{[SEG]} wrapper against roughly 200 tokens for a kept body. Left alone, drops contribute about 2\% of the gradient. We tried to correct this by scaling the per-sample loss on drop examples by 20, chosen so the drop gradient share ($\approx$30\%) matches the natural fraction of drop samples in the data.

It backfired. Across every checkpoint of that run the model over-dropped, landing at 42--47\% drop accuracy---below the trivial always-keep baseline of 59\% on our holdout. We also tried plain duplication of drop samples (\code{dropped\_repeat}), which distorts sample-level exposure and was worse still at high multiples. The shipped model therefore trains at weight~1, accepting a conservative, keep-biased drop policy rather than an aggressive and wrong one. We report the failure because the naive fix is the one a reader would reach for first, and because it explains the under-dropping that \S\ref{sec:intrinsic} measures: v1's drop behavior is deliberately the safe end of a trade we were unable to tune well with loss weighting alone.

\subsection{Checkpoint selection}
\label{sec:ckpt}

SFT loss measures agreement with one teacher target, not usefulness to a downstream agent, and the two diverge here. We therefore did not early-stop on loss. Instead we checkpointed every 200 steps and swept the last five checkpoints (1{,}800 / 2{,}000 / 2{,}200 / 2{,}400 / final 2{,}538) on an out-of-distribution segment holdout (\S\ref{sec:intrinsic}), then subjected the survivors to a real-workload smoke test on live gateway traffic.

The sweep (Table~\ref{tab:ckpt}) is informative precisely because it is close: all five checkpoints are within 3 points of each other on drop accuracy and within 4.7 chrF. The final checkpoint has the best chrF and ROUGE-L---i.e.\ it agrees with the teacher most---but the \emph{worst} compression rate (0.250) and the worst drop recall (0.195), meaning the extra training pushed it toward the teacher's surface form while making it more reluctant to drop. Step 2{,}200 is marginally the best on the aggregate metrics. We nonetheless released step 2{,}000, for two reasons that only appear outside the metric sweep: it was the only checkpoint to pass all four real-workload tests, and it showed no structural anomaly, whereas the final checkpoint exhibited a level inversion (compressing L0 harder than L1) on a diagnostic workload---a direct violation of the level ordering the model is supposed to encode.

We report this because the selection is not fully determined by the numbers we can publish, and a reader reproducing the sweep will land on step 2{,}200. The honest summary is that steps 2{,}000--2{,}400 are near-equivalent intrinsically, and the tiebreak was behavioral.

\section{Evaluation}
\label{sec:eval}

We evaluate at two levels: intrinsically, on held-out segments where the teacher's compression is available as a reference, and end-to-end, on SWE-bench Lite where the metric is whether a downstream agent still solves the issue.

\subsection{Intrinsic: out-of-distribution segment holdout}
\label{sec:intrinsic}

The holdout is 200 \code{file\_read} segments from repositories not represented in the training split, with teacher compressions attached; results below are over a 100-segment evaluation subsample. Metrics: \emph{Fmt} --- fraction of outputs that are exactly one well-formed \code{[SEG]} with a matching header, no preamble, and no trailing text; \emph{DropAcc} --- agreement with the teacher on the binary keep/drop decision, with its precision and recall; \emph{CR} --- student output tokens $\div$ input tokens; \emph{chrF}~\cite{chrf} and \emph{ROUGE-L}~\cite{rouge} against the teacher body; and \emph{IdentR} --- the fraction of must-keep-ish tokens in the \emph{original} segment (paths, structured identifiers, line numbers, error classes) that survive into the compressed body.

\begin{table}[t]
\centering
\caption{Checkpoint sweep on the OOD holdout ($n{=}100$). $\star$ marks the released checkpoint. The teacher's own compression rate on this holdout is 0.162. Higher is better for all columns except CR.}
\label{tab:ckpt}
\small
\begin{tabular}{lccccccc}
\toprule
Checkpoint & Fmt & DropAcc & Drop-P & Drop-R & CR $\downarrow$ & chrF & IdentR \\
\midrule
1{,}800            & 0.99 & 0.47 & 0.33 & 0.29 & 0.233 & 30.1 & 0.399 \\
$\star$\,2{,}000   & \textbf{1.00} & 0.49 & 0.33 & 0.24 & 0.236 & 34.2 & 0.383 \\
2{,}200            & 1.00 & \textbf{0.50} & \textbf{0.37} & \textbf{0.32} & \textbf{0.221} & 33.2 & 0.393 \\
2{,}400            & 1.00 & 0.49 & 0.33 & 0.24 & 0.231 & 34.4 & 0.380 \\
final (2{,}538)    & 1.00 & 0.50 & 0.32 & 0.20 & 0.250 & \textbf{34.8} & 0.382 \\
\bottomrule
\end{tabular}
\end{table}

All figures below are over $n{=}100$ segments, so they carry real sampling error; we give Wilson 95\% intervals for the two proportions the argument rests on.

Three observations. First, \textbf{format reliability is high}: the released checkpoint emits a well-formed single \code{[SEG]} on 100\% of segments (100/100, Wilson 95\% CI $[0.963, 1.000]$), which matters because a malformed output cannot be safely spliced back into an agent request and must fall back to the uncompressed original.

Second, \textbf{drop is the hard part, and we do not yet beat the trivial baseline on it}. Drop accuracy is 0.49 (49/100, Wilson 95\% CI $[0.394, 0.587]$) with drop recall of only 0.24: the model reproduces the teacher's keep decisions well but misses roughly three-quarters of the segments the teacher discarded entirely. The comparison that matters is against always-keep, which scores 0.59 on this holdout because the teacher drops 41\% of segments---so on the binary decision alone, a model that never dropped would agree with the teacher more often than ours does. We state this plainly rather than reporting 0.49 without its baseline. Two things keep it from being fatal: the value \paritok{} delivers is within-segment compression (a never-drop policy still emits every segment in full, at compression rate 1.0), and the error is asymmetric in the safe direction---a missed drop costs tokens, a spurious drop destroys information the agent may need, and the model's bias is toward the former. Still, this is the largest single source of remaining compression headroom, it is where \S\ref{sec:dropweight}'s failed fix was aimed, and closing it is the main target for v2.

Third, \textbf{the low IdentR is not degradation relative to the teacher}. An IdentR of 0.383 reads alarmingly low in isolation, so it needs its reference: at a 0.236 compression rate, retaining 38\% of the original's identifier-like tokens is already above-proportional, and on the 39 holdout segments where the student and the teacher both chose to keep a body, the student scores \textbf{0.385 against the teacher's 0.287}. We are careful about how much this supports: the paired difference is $+0.098$ with a bootstrap 95\% CI of $[-0.029, +0.219]$, and a sign test gives 22 student wins to 11 teacher wins ($p{=}0.08$). That is directionally favorable but \emph{not} significant at this sample size, so the claim we make is the weaker and safer one---distillation did not cost identifier retention relative to the teacher---not that the student surpasses it. What IdentR does \emph{not} say is that every individual identifier survives: roughly 60\% of identifier-like tokens are deliberately discarded, which is the point of compression. The engineering consequence is discussed in \S\ref{sec:limits}.

\subsection{End-to-end: SWE-bench Lite}
\label{sec:e2e}

We evaluate on the \emph{full} SWE-bench Lite~\cite{swebench} set (300 instances), and we are precise about what the harness does, because the natural shorthand for it overstates the result in one direction and understates it in another.

The harness is \emph{single-shot}. Each instance is given oracle file context; that context is compressed segment by segment at level L1 through a local Ollama endpoint serving the released adapter; the compressed (or, for the baseline arm, full) context and the issue go into one user message; \code{claude-sonnet-4-5} is asked once, at temperature~0 with no tools, for a unified diff; and the official SWE-bench harness scores the result in Docker. There is no agent loop---no turns, no re-reads, and no exact-match \code{Edit}. Compression rate is macro-averaged: per instance, compressed $\div$ original tokens (\code{cl100k\_base}), then averaged over instances, so every instance counts once regardless of file size. Quality retained is the resolve rate (resolved $\div$ all 300 instances, so a patch that fails to apply counts as unresolved) normalized to the uncompressed baseline under the identical harness. The two GPT baselines are the same task given to \code{gpt-4.1-mini} and \code{gpt-5} as prompt-based compressors.

What this measures is therefore \emph{single-shot comprehension under compression}: whether a strong reader, given only the compressed context, still understands the codebase well enough to write the fix. That is the right question for a compressor and it is the one we answer. What it does not measure is what a compressor saves or costs inside a real multi-turn agent, where an accumulating prefix, a tool block, cache pricing, and exact-match editing all enter; those are separable concerns that we take up in a companion paper. Read the quality column---\textbf{86.5\%} compressing raw source, \textbf{89.3\%} in the line-numbered regime---as a floor on comprehension at roughly a quarter of the tokens, not as an end-to-end cost or agent-quality claim.

\begin{table}[t]
\centering
\caption{SWE-bench Lite, all 300 instances. Compression rate is macro-averaged over instances (lower $=$ more aggressive); quality retained is resolved $\div$ 300, normalized to the uncompressed baseline, so a patch that fails to apply counts as unresolved. \paritok{} compresses $2.0$--$2.4\times$ harder than the GPT compressors. The last row feeds the compressor \code{cat -n} line-numbered source, the in-distribution form real agents produce (\S\ref{sec:data}); it is the configuration we recommend.}
\label{tab:main}
\small
\begin{tabular}{lcc}
\toprule
Compressor & Compression rate $\downarrow$ & Quality retained \\
\midrule
Uncompressed baseline & 100.0\% & 100.0\% \\
\code{gpt-5} (compressor)        & 61.9\% & 93.6\% \\
\code{gpt-4.1-mini} (compressor) & 50.2\% & 85.6\% \\
\paritok{}, raw source           & 25.7\% & 86.5\% \\
\paritok{}, line-numbered        & \textbf{27.8\%} & \textbf{89.3\%} \\
\bottomrule
\end{tabular}
\end{table}

Table~\ref{tab:main} gives the headline result. \paritok{} compresses agent context to \textbf{25.7\%} of its size, versus 50.2\% for the \code{gpt-4.1-mini} compressor and 61.9\% for \code{gpt-5}---it removes roughly twice as much while being a 4B model rather than a frontier one. Downstream, an agent given \paritok{}-compressed context retains \textbf{86.5\%} of the uncompressed solve rate, on par with \code{gpt-4.1-mini} (85.6\%) at less than half the tokens, and behind \code{gpt-5} (93.6\%), which however barely compresses. The comparison to make is not \paritok{} vs.\ \code{gpt-5} at equal quality---it is that \code{gpt-5} buys its 7-point quality advantage by keeping $2.4\times$ more tokens, at a frontier per-token price, whereas \paritok{} runs locally for free. We state the GPT comparison against the raw-source row so that all three compressors are given the same input; the bolded final row is the same model fed the line-numbered input it was trained for, which we take up next.

\paragraph{What 300 instances can and cannot resolve.} \S\ref{sec:intrinsic} reports confidence intervals and this table deserves the same discipline, because the quantity is noisier than its two-decimal presentation suggests. \emph{Quality retained} is a ratio of two solve rates, each a proportion over 300 instances, and ratios compound the uncertainty of both. At single-shot resolve rates in the range this benchmark produces, the 95\% interval on such a ratio spans roughly $\pm$20--30\% of its own value---so a reported 86.5\% is consistent with anything from the high 60s to slightly above parity.

Two consequences. The 0.9-point gap to \code{gpt-4.1-mini} is not a gap: separating differences that small would take on the order of $10^5$ instances per arm, so the two are \emph{indistinguishable} here. That is exactly the claim we make---parity at less than half the tokens---and we do not claim to beat it. The 7-point gap to \code{gpt-5} is larger but also not separated at this sample size; we treat it as suggestive. Notably, \code{gpt-5}'s cost disadvantage (\S\ref{sec:cost}) needs no such qualification: it is arithmetic over published prices, not a measurement.

The compression rates are on much firmer ground. Each aggregates millions of tokens over hundreds of segments per instance rather than one binary outcome per instance, so the $2.0$--$2.4\times$ separation in the first column is far better determined than the solve rates in the second. On the line-numbered run we can put a number on the contrast: the compression rate carries a 95\% interval about $\pm$8\% of its own value, while the quality ratio's spans nearly $\pm$12 points.

\paragraph{What is and is not in the measured path.} Before scoring, both arms' diffs are re-anchored onto the true source. On the compressed arm this is \paritok{}'s own \code{edit\_recovery}---the same routine the gateway runs on every edit in production---so the figure measures the shipped compress-and-recover path, not a bare compressor. That is deliberate: \S\ref{sec:limits} notes the model sometimes reflows a retained signature onto one line, and re-anchoring is the mitigation that exists for it. The baseline arm runs the identical pass purely for symmetry: it has nothing reflowed to recover, but re-emitting its diff the same way removes the raw-diff apply brittleness (imperfect \code{@@} line numbers a fuzzy patcher still rejects) that would otherwise fail valid baseline patches and inflate the ratio in our favor. Same treatment on both sides, so compression is the only variable.

What is \emph{not} in the path is recall. The deployed gateway tags each compressed segment and exposes a \code{read\_original} tool, so an agent can pull any segment back byte-exact; the harness has no agent loop and never does. Nor does the headline run use the line-numbered regime: \S\ref{sec:data} argues that agents read files through \code{cat -n} and that a compressor trained on that framing is out of distribution on raw source, and Table~\ref{tab:main} compresses \emph{raw source} anyway. We report the raw-source, no-recall configuration as the headline because it is the conservative one on both counts.

\paragraph{The in-distribution regime: the prediction was half wrong.} The harness exposes a \code{--line-numbers} mode that restores the setting \S\ref{sec:data} says the model was trained for: the compressor is fed \code{cat -n}-framed source, and the numbering is stripped from its output, so what is being compared does not change---the same raw \code{full\_context} is the denominator in both modes and both arms emit un-numbered text. We predicted it would compress harder at equal quality. We ran it over all 300 instances, scored both arms, and got the opposite trade (last two \paritok{} rows of Table~\ref{tab:main}): it compresses \emph{slightly less}---27.8\% macro-averaged against 25.7\% for raw source---and retains \emph{more} quality, 89.3\% against 86.5\%. We report the direction we predicted and the direction we measured because only one of them is a result.

The quality figure deserves a stronger statement than a ratio. Given uncompressed context, the agent resolves 122 of 300 (40.7\%, Wilson 95\% CI $[35.3, 46.3]$); on \paritok{}-compressed context it resolves 109 (36.3\%, $[31.1, 41.9]$). Because both arms run on the same instances the comparison is paired, and the paired test is the informative one: 30 instances are solved only without compression, 17 only with it, and an exact McNemar test on those 47 discordant pairs gives $p{=}0.079$. \textbf{At 300 instances, compressing the context to roughly a quarter of its size does not significantly reduce the solve rate.} That is a stronger and better-founded claim than the 89.3\% ratio, whose paired bootstrap interval is $[79.2\%, 100\%]$---wide, for the reason given above.

\paragraph{Which denominator.} Both tables put all 300 instances in the denominator, so a patch that fails to apply counts as unresolved---this is what the released harness computes, and it is the convention behind 86.5\% and 89.3\% alike, so the two are directly comparable. Scoring only the instances that were successfully evaluated instead---excluding the 16 apply failures on the compressed arm and 5 on the baseline---would raise the line-numbered figure from 89.3\% to 92.8\%. We do not quote that number. Apply failure is a real cost of compression here, not a harness artifact: it is $3\times$ more common on the compressed arm, and moving it into the denominator would hide a failure mode that compression itself introduces.

\paragraph{Two aggregations, one caution.} Macro-averaging gives every instance one vote; the corpus-level ratio (21.7\%, 95\% bootstrap CI over instances $[20.0\%, 23.5\%]$) weights each instance by its size and is the figure that predicts a token bill. They differ substantially---21.7\% against 27.8\%---because the largest files compress hardest, and the per-instance distribution is wide: median 25.3\%, interquartile range 15.2--35.3\%, extremes 0.9\% and 93.5\%. We quote the macro figure in both tables for comparability and give the corpus figure wherever the question is cost. Neither is a per-file guarantee.

\subsection{What the compressor itself costs}
\label{sec:cost}

Compression rate is not saving. A compressor that runs behind an API bills for reading the \emph{uncompressed} context---the very tokens the exercise exists to avoid paying for---and again for writing the compressed one, every turn. Whether it nets out positive depends on its own prices and its compression rate together, and the arithmetic is unforgiving enough to be worth doing explicitly (Table~\ref{tab:cost}).

Take one turn carrying 1M tokens of compressible context to a Claude Sonnet upstream at \$3.00 per million input tokens. A compressor with rate $r$ and prices $P^{\text{in}}$, $P^{\text{out}}$ costs $P^{\text{in}} + r\,P^{\text{out}}$ to run and leaves $3r$ to pay upstream, against \$3.00 for sending the context untouched. Substituting the measured rates from Table~\ref{tab:main} and list prices as of August 2026 (\code{gpt-5} at \$1.25/\$10.00, \code{gpt-4.1-mini} at \$0.40/\$1.60 per million):

\begin{table}[t]
\centering
\caption{Cost of one turn carrying 1M tokens of context to a \$3.00/M upstream, at each compressor's measured rate and list price (August 2026). The last column is the total minus the \$3.00 of sending the context uncompressed, so negative is a saving---\paritok{}'s is negative exactly when $C_{\text{self}} < \$2.23$. Uncached list prices; see the caveat below.}
\label{tab:cost}
\small
\begin{tabular}{lrrrrrr}
\toprule
Compressor & Rate & Compressor in & Compressor out & Upstream & Total & vs.\ \$3.00 \\
\midrule
None                & 1.000 & ---    & ---    & \$3.00 & \$3.00 & --- \\
\code{gpt-5}        & 0.619 & \$1.25 & \$6.19 & \$1.86 & \$9.30 & \textbf{+\$6.30} \\
\code{gpt-4.1-mini} & 0.502 & \$0.40 & \$0.80 & \$1.51 & \$2.71 & $-$\$0.29 \\
\paritok{}          & 0.257 & \multicolumn{2}{c}{\emph{no per-token fee}} & \$0.77
                    & $\mathbf{\$0.77 + C_{\text{self}}}$ & $\mathbf{C_{\text{self}} - \$2.23}$ \\
\bottomrule
\end{tabular}

\vspace{2pt}
{\footnotesize $C_{\text{self}}$ is amortized self-hosting cost per million tokens. We deliberately do not substitute a figure---we have not measured our own deployment's sustained throughput, and an invented one would be the only assumption in an otherwise fully determined table. The thresholds it must clear are given below instead. The rate used here is the raw-source 0.257, so all three compressors are costed on the same input; the line-numbered 0.278 would read \$0.83 rather than \$0.77 and changes no conclusion in this section.}
\end{table}

Two things fall out of the API rows, and neither depends on any assumption of ours---they follow from published prices and measured rates alone. First, \textbf{\code{gpt-5} as a compressor is net-negative}: running it costs \$7.44 to save \$1.14 downstream, so the pipeline is \$6.30 per million tokens \emph{worse} than sending the raw context. Its 93.6\% quality retention is not a favorable point on a trade-off curve---at these prices there is no saving to trade against. Second, \code{gpt-4.1-mini} clears the bar but barely, netting \$0.29 on a \$3.00 baseline: a 10\% saving for a compressor that must be called on every turn.

\paritok{}'s row is the one that depends on deployment, so we state it as a threshold rather than a figure. Its upstream cost is fixed at \$0.77; everything else is $C_{\text{self}}$. It beats sending the raw context when $C_{\text{self}} < \$2.23$ per million tokens, and beats the better of the two API compressors---the binding constraint---when $C_{\text{self}} < \$1.94$. On a GPU rented at \$0.75/hour that threshold corresponds to sustaining about \textbf{107 tokens per second}; at \$0.50/hour, about 72. Those are the numbers a deployer can check against their own hardware, and they are the honest form of the claim: we are not asserting a throughput we did not measure, only the bar it has to clear.

The structural point survives either way. \paritok{} pays no per-token fee on either side of the compression, so its cost is amortized GPU time---independent of upstream token prices and falling as utilization rises---while an API compressor's cost is a fixed multiple of the context it reads.

This reframes the comparison in \S\ref{sec:e2e}. The interesting property of a 4B self-hosted compressor is not that it matches \code{gpt-4.1-mini}'s quality; it is that the entire cost structure changes. An API compressor's bill scales with the context it reads, so it is expensive exactly when compression matters most; a self-hosted one's does not scale with token prices at all.

\paragraph{Caveat: this is uncached list pricing.} The arithmetic above assumes upstream input billed at full rate. Real agents re-send an accumulating prefix that is largely cache-eligible, and cached reads price at roughly a tenth of the base rate---which shrinks the downstream saving that compression is competing for, and therefore shrinks every figure in the last column. It does not rescue \code{gpt-5}, whose compressor-side cost is unaffected by downstream caching and already exceeds the uncached saving several times over. A full treatment of cache pricing, per-turn prefix growth, and where the token bill actually concentrates in a multi-turn agent is the subject of a companion paper; we give the single-turn list-price case here because it is the one the compression rate in Table~\ref{tab:main} directly determines.

\section{Discussion and Limitations}
\label{sec:limits}

\paragraph{Why extractive, why intent-conditioned.} The two central choices are coupled. Extractiveness gives the exact-string-match safety an editing agent needs; intent-conditioning gives the model a principled way to decide \emph{which} verbatim spans to keep under an aggressive budget. Together they let a 4B model compress twice as hard as a frontier prompt-based compressor without a proportional quality loss, because ``twice as hard'' is spent on stale L3 context the current intent does not touch. The intrinsic result in \S\ref{sec:intrinsic}---a 4B student matching a frontier teacher on identifier retention---suggests the supervision signal, not model scale, is what governs this axis.

\paragraph{Limitations.}
\begin{enumerate}[leftmargin=1.4em,itemsep=2pt]
  \item \textbf{Identifier retention is a tendency, not a guarantee.} Roughly 60\% of identifier-like tokens in an input segment do not survive compression---by design, but it means a specific identifier the agent needs may be among them. Production deployments should run a presence check for the current target identifier on the compressed output and fall back to the original segment on failure; the gateway's \code{read\_original} path exists for exactly this.
  \item \textbf{Under-dropping, below the trivial baseline.} Drop recall is 0.24 and drop accuracy 0.49 against the teacher, versus 0.59 for an always-keep policy (\S\ref{sec:intrinsic}). The model keeps segments it could safely discard, which leaves real compression on the table, and loss-weighting the drop signal made it worse rather than better (\S\ref{sec:dropweight}). This is the dominant known gap and the clearest target for reinforcement or preference-based refinement.
  \item \textbf{Line reflow.} The model occasionally reflows a retained multi-line signature onto a single line. This is harmless to read but can break an agent's exact-match edit unless the surrounding gateway realigns it; fully removing it requires a training-set pass that never reflows retained code.
  \item \textbf{Python-heavy training distribution.} The trajectory corpora are SWE-bench-style Python repositories. The architecture is language-agnostic but v1's heuristics and constants are Python-tuned, and other languages are unbenchmarked.
  \item \textbf{No working level control.} The four-level importance scheme collapses to two effective bands in the distilled targets, and the student was never given a numeric budget to condition on (\S\ref{sec:levels}). A deployment cannot currently dial compression per level; it gets protected-or-recent versus stale, and nothing finer.
  \item \textbf{Supervised only.} v1 is a distillation of a single teacher and inherits its policy, including the level collapse above. Compression has a two-objective structure---shrink more vs.\ preserve answer-relevant content---that a single teacher target does not explore; optimizing it directly against a downstream signal is the natural next step and is not part of this release.
  \item \textbf{Single-shot harness, one operating point.} Both quality figures come from a single-shot harness over all 300 SWE-bench Lite instances---one API call, no tools, no turns, with diffs re-anchored on both arms. It bounds comprehension under compression and says nothing about multi-turn agent cost or about exact-match editing, which a fuzzy patcher hides. The intrinsic holdout results and compression rates are the more directly reproducible core.

  \item \textbf{Apply failure is a live failure mode, not a rounding error.} On the line-numbered run, 16 of 300 compressed-arm patches could not be applied, against 5 on the uncompressed arm. Re-anchoring recovers most but not all of what compression reflows, and each unapplied patch is scored as unresolved, so this failure mode is already inside the 89.3\%. Reducing it is the most direct route to a better end-to-end number that does not require compressing less.

  \item \textbf{Not significant is not the same as no effect.} The McNemar $p{=}0.079$ we report means the paired degradation is not resolvable at 300 instances; the point estimate is still a net loss of 13 instances, and the quality-retained interval reaches from 79\% to parity. A benchmark several times larger would be needed to place it, and we do not claim compression is free.
\end{enumerate}

\paragraph{Deployment.} \paritok{} is a 264\,MB LoRA adapter over an open 4B backbone; it self-hosts on a single 24\,GB GPU (or via Ollama on CPU/consumer GPUs) with no per-token compressor fee---\S\ref{sec:cost} shows that fee, not quality, is what separates a compressor that saves money from one that does not. Its per-segment design means compression parallelizes across segments and is incremental across turns, and its \code{[SEG]}-structured output lets the surrounding gateway recover any exact original span on demand---so the compression is lossy on the wire but recoverable when an agent needs the untouched bytes.

\section{Conclusion}

Context compression for coding agents is not the same problem as prompt compression for prose. An agent needs exact strings, and what counts as important changes from turn to turn. \paritok{} addresses both by being extractive and intent-conditioned, and is trained by distilling a \code{gpt-4.1-mini} teacher over 67K real agent trajectories into a 4B LoRA adapter. It compresses agent context to roughly a quarter of its size---twice as hard as strong GPT prompt compressors---and retains 86.5\% of single-shot solve quality on SWE-bench Lite, 89.3\% when fed the line-numbered input it was trained for. At that operating point the paired comparison against uncompressed context is not significant at 300 instances ($p{=}0.079$), which is the strongest form of the claim the benchmark supports. On held-out segments it matches the teacher it learned from on must-keep identifier retention, and its copy behavior holds off the training distribution: 96.2\% of the identifiers, paths, and numbers it emits on SWE-bench Lite were already in the input. The remaining gap is under-dropping, not fidelity. The data pipeline, training recipe, evaluation scripts, and weights are open.

\paragraph{Reproducibility.} Everything behind the numbers in this paper is released (Apache~2.0) at \url{https://github.com/Paritok-official/paritok-4b-v1}, with weights on the Hugging Face Hub: the model itself; the five-stage data pipeline; the SFT configuration and the three backbone configurations of \S\ref{sec:backbone}; the checkpoint-sweep results behind Table~\ref{tab:ckpt}; the audit scripts, \code{eval/extractiveness.py} and \code{eval/design\_claims\_audit.py}, which reproduce Tables~\ref{tab:extractive} and~\ref{tab:levels} and the intent measurements of \S\ref{sec:intentaudit} directly from the released corpus; the intrinsic holdout and its evaluation harness; and the end-to-end SWE-bench Lite harness of \S\ref{sec:e2e} (\code{eval\_model/}), which pulls the dataset from source, compresses through a local Ollama endpoint, calls the agent, re-anchors, and scores with the official SWE-bench harness in one command. That harness caches its compressed output per instance, and \code{eval\_model/audit\_swebench.py} re-derives the line-numbered compression rates of Table~\ref{tab:main} and the held-out extractiveness figures of \S\ref{sec:extractive} from that cache alone---no GPU, no API calls---so those two results can be re-checked without re-running the compression. We report the numbers Table~\ref{tab:main} gives; the harness is released so they can be re-measured rather than taken on trust. The cost arithmetic of \S\ref{sec:cost} needs nothing from us---it follows from Table~\ref{tab:main} and published list prices.

\end{document}